%% file: main.tex
\documentclass[10pt,twocolumn,letterpaper]{article}

\pdfoutput=1
\usepackage{cvpr}
\usepackage{times}
\usepackage{epsfig}
\usepackage{graphicx}
\usepackage{amsmath}
\usepackage{amssymb}

\usepackage{amsfonts}
\usepackage{color}
\usepackage{caption}

\usepackage[pagebackref=true,breaklinks=true,letterpaper=true,colorlinks,bookmarks=false]{hyperref}

\cvprfinalcopy 

\def\cvprPaperID{932} 

\ifcvprfinal\fi
\begin{document}

\title{Inverse Compositional Spatial Transformer Networks}

\author{Chen-Hsuan Lin \qquad Simon Lucey\\
The Robotics Institute \\ Carnegie Mellon University\\
{\tt\small chenhsul@andrew.cmu.edu \qquad slucey@cs.cmu.edu}
}

\maketitle

\input{notation}
\definecolor{CNNcolor}{rgb}{0,0.3,0.6}
\definecolor{STcolor}{rgb}{0.2,0.4,0}
\newcommand{\red}{\color{red}}
\newcommand{\cyan}{\color{cyan}}
\newcommand{\black}{\color{black}}
\newcommand{\CNNcolor}{\color{CNNcolor}}
\newcommand{\STcolor}{\color{STcolor}}

\newcommand\numberthis{\addtocounter{equation}{1}\tag{\theequation}}

\begin{abstract}

In this paper, we establish a theoretical connection between the classical Lucas \& Kanade (LK) algorithm and the emerging topic of Spatial Transformer Networks (STNs).
STNs are of interest to the vision and learning communities due to their natural ability to combine alignment and classification within the same theoretical framework.
Inspired by the Inverse Compositional (IC) variant of the LK algorithm, we present Inverse Compositional Spatial Transformer Networks (IC-STNs).
We demonstrate that IC-STNs can achieve better performance than conventional STNs with less model capacity; in particular, we show superior performance in pure image alignment tasks as well as joint alignment/classification problems on real-world problems. 


\end{abstract}

\section{Introduction}

Recent rapid advances in deep learning are allowing for the learning of complex functions through convolutional neural networks (CNNs), which have achieved state-of-the-art performances in a plethora of computer vision tasks~\cite{krizhevsky2012imagenet,simonyan2014very,he2015deep}. Most networks learn to tolerate spatial variations through: (a) spatial pooling layers and/or (b) data augmentation techniques~\cite{simard2003best}; however, these approaches come with several drawbacks. Data augmentation (\ie the synthetic generation of new training samples through geometric distortion according to a known noise model) is probably the oldest and best known strategy for increasing spatial tolerance within a visual learning system. This is problematic as it can often require an exponential increase in the number of training samples and thus the capacity of the model to be learned. Spatial pooling operations can partially alleviate this problem as they naturally encode spatial invariance within the network architecture and uses sub-sampling to reduce the capacity of the model. However, they have an intrinsic limited range of tolerance to geometric variation they can provide; furthermore, such pooling operations destroy spatial details within the images that could be crucial to the performance of subsequent tasks. 

Instead of designing a network to solely give tolerance to spatial variation, another option is to have the network solve for some of the geometric misalignment in the input images~\cite{NIPS2014_5420,NIPS2012_4769}. Such a strategy only makes sense, however, if it has lower capacity and computational cost as well as better performance than traditional spatially invariant CNNs. Spatial Transformer Networks (STNs)~\cite{jaderberg2015spatial} are one of the first notable attempts to integrate low capacity and computationally efficient strategies for \textit{resolving} - instead of \textit{tolerating} - misalignment with classical CNNs. Jaderberg \etal presented a novel strategy for integrating image warping within a neural network and showed that such operations are (sub-)differentiable, allowing for the application of canonical backpropagation to an image warping framework. 

The problem of learning a low-capacity relationship between image appearance and geometric distortion is not new in computer vision. Over three and a half decades ago, Lucas \& Kanade (LK)~\cite{lucas1981iterative} proposed the seminal algorithm for gradient descent image alignment. The LK algorithm can be interpreted as a feed forward network of multiple alignment modules; specifically, each alignment module contains a low-capacity predictor (typically linear) for predicting geometric distortion from relative image appearance, followed by an image resampling/warp operation. The LK algorithm differs fundamentally, however, to STNs in their application: image/object alignment instead of classification. 

Putting applications to one side, the LK and STN frameworks share quite similar characteristics however with a criticial exception. In an STN with multiple feed-forward alignment modules, the output image of the previous alignment module is directly fed into the next. As we will demonstate in this paper, this is problematic as it can create unwanted boundary effects as the number of geometric prediction layers increase.
The LK algorithm does not suffer from such problems; instead, it feeds the warp parameters through the network (instead of the warped image) such that each subsequent alignment module in the network resamples the original input source image.
Furthermore, the Inverse Compositional (IC) variant of the LK algorithm~\cite{baker2004lucas} has demonstrated to achieve equivalently effective alignment by reusing the same geometric predictor in a compositional update form. 

Inspired by the IC-LK algorithm, we advocate an improved extension to the STN framework that (a) propagates \textit{warp parameters}, rather than \textit{image intensities}, through the network, and (b) employs the same geometric predictor that could be reapplied for all alignment modules. We propose Inverse Compositional Spatial Transformer Networks (IC-STNs) and show its superior performance over the original STNs across a myriad of tasks, including pure image alignment and joint alignment/classification problems. 

We organize the paper as follows:
we give a general review of efficient image/object alignment in Sec.~\ref{sec:alignment} and an overview of Spatial Transformer Networks in Sec.~\ref{sec:STN}. We describe our proposed IC-STNs in detail in Sec.~\ref{sec:ICSTN} and show experimental results for different applications in Sec.~\ref{sec:experiments}. Finally, we draw to our conclusion in Sec.~\ref{sec:conclusion}.

\section{Efficient Image \& Object Alignment} \label{sec:alignment}

In this section, we give a review of nominal approaches to efficient and low-capacity image/object alignment.

\subsection{The Lucas \& Kanade Algorithm}

The Lucas \& Kanade (LK) algorithm~\cite{lucas1981iterative} has been a popular approach for tackling dense alignment problems for images and objects. For a given geometric warp function parameterized by the warp parameters $\p$, one can express the LK algorithm as minimizing
the sum of squared differences (SSD) objective in the image space,
\begin{equation} \label{eq:LK}
\min_{\deltap} \eucsqnorm{ \I(\p+\deltap) - \T(\0) } ,
\end{equation}
where $\I$ is the source image, $\T$ is the template image to align against, and $\deltap$ is the warp update being estimated. Here, we denote $\I(\p)$ as the image $\I$ warped with the parameters $\p$.
The LK algorithm assumes a approximate linear relationship between appearance and geometric displacements; specifically, it linearizes~\eqref{eq:LK} by taking the first-order Taylor approximation as
\begin{equation} \label{eq:LK-linearize}
\min_{\deltap} \eucsqnorm{ \I(\p) + \frac{\partial\I(\p)}{\partial\p}\deltap - \T(\0) } .
\end{equation}
The warp parameters are thus additively updated through $\p \leftarrow \p+ \deltap$, which can be regarded as a quasi-Newton update.
The term $\frac{\partial\I(\p)}{\partial\p}$, known as the steepest descent image, is the composition of image gradients and the predefined warp Jacobian, where the image gradients are typically estimated through finite differences.
As the true relationship between appearance and geometry is seldom linear, the warp update $\deltap$ must be iteratively estimated and applied until convergence is reached.

A fundamental problem with the canonical LK formulation, which employs addtive updates of the warp parameters, is that $\frac{\partial\I(\p)}{\partial\p}$ must be recomputed on the rewarped images for each iteration, greatly impacting computational efficiency. 
Baker and Matthews~\cite{baker2004lucas} devised a computationally efficient variant of the LK algorithm, which they referred to as the Inverse Compositional (IC) algorithm. 
The IC-LK algorithm reformulates~\eqref{eq:LK} to predict the warp update to the template image instead, written as
\begin{equation} \label{eq:LKIC}
\min_{\deltap} \eucsqnorm{ \I(\p) - \T(\deltap) } ,
\end{equation}
and the linearized least-squares objective is thus formed as
\begin{equation} \label{eq:LKIC-linearize}
\min_{\deltap} \eucsqnorm{ \I(\p) - \T(\0) - \frac{\partial\T(\0)}{\partial\p}\deltap } .
\end{equation}
The least-squares solution is given by
\begin{equation} \label{eq:LKIC-solution}
\deltap = \left( \frac{\partial\T(\0)}{\partial\p} \right)^{\dagger} \left( \I(\p)-\T(\0) \right) ,
\end{equation}
where the superscript $\dagger$ denotes the Moore-Penrose pseudo-inverse operator. This is followed by the inverse compositional update $\p \leftarrow \p \circ (\deltap)^{-1}$, where we abbreviate the notation $\circ$ to be the composition of warp functions parameterized by $\p$, and $(\deltap)^{-1}$ is the parameters of the inverse warp function parameterized by $\deltap$. 

The solutions of~\eqref{eq:LK-linearize} and~\eqref{eq:LKIC-linearize} are in the form of linear regression, which can be more generically expressed as 
\begin{equation} \label{eq:app-geo-linear}
\deltap = \R \cdot \I(\p) +\b ,
\end{equation}
where $\R$ is a linear regressor establishing the linear relationship between appearance and geometry, and $\b$ is the bias term.
Therefore, LK and IC-LK can be interpreted as belonging to the category of cascaded linear regression approaches for image alignment.

It has been shown~\cite{baker2004lucas} that the IC form of LK is effectively equivalent to the original form; the advantage of the IC form lies in its efficiency of computing the fixed steepest descent image $\frac{\partial\T(\0)}{\partial\p}$ in the least-squares objective. Specifically, it is evaluated on the static template image $\T$ at the identity warp $\p=\0$ and remains constant across iterations, and thus so is the resulting linear regressor $\R$. This gives an important theoretical proof of concept that a fixed predictor of geometric updates can be successfully employed within an iterative image/object alignment strategy, further reducing unnecessary model capacities.

\subsection{Learning Alignment from Data}

More generally, cascaded regression approaches for alignment can be learned from data given that the distribution of warp displacements is known a priori. A notable example of this kind of approach is the Supervised Descent Method (SDM)~\cite{xiong2013supervised}, which aims to learn the series of linear geometric predictors $\{\R,\b\}$ from data. The formulation of SDM's learning objective is
\begin{equation} \label{eq:SDM}
\min_{\R,\b} \sum_{n=1}^{N} \sum_{j=1}^{M} \eucsqnorm{ \delta\p_{n,j} - \R \cdot \I_n(\p_n \circ \delta\p_{n,j}) -\b } ,
\end{equation}
where $\delta\p$ is the geometric displacement drawn from a known generating distribution using Monte Carlo sampling, and $M$ is the number of synthetically created examples for each image.
Here, the image appearance $\I$ is often replaced with a predefined feature extraction function (\eg SIFT~\cite{lowe2004distinctive} or HOG~\cite{dalal2005histograms}) of the image.
This least-squares objective is typically solved with added regularization (\eg ridge regression) to ensure good matrix condition.

SDM is learned in a sequential manner, \ie the training data for learning the next linear model is drawn from the same generating distribution and applied through the previously learned regressors.
This has been a popular approach for its simplicity and effectiveness across various alignment tasks, leading to a large number of variants~\cite{ren2014face,asthana2014incremental,lin2016conditional} of similar frameworks.
Like the LK and IC-LK algorithms, SDM is another example of employing multiple low-capacity models to establish the nonlinear relationship between appearance and geometry. We draw the readers' attention to~\cite{lin2016conditional} for a more formally established link between LK and SDM.

It is a widely agreed that computer vision problems can be solved much more efficiently if misalignment among data is eliminated. Although SDM learns alignment from data and guarantees optimal solutions after each applied linear model, it is not clear whether such alignment learned in a greedy fashion is optimal for the subsequent tasks at hand, \eg classification. In order to optimize in terms of the \textit{final} objective, it would be more favorable to paramterize the model as a deep neural network and optimize the entire model using backpropagation.


\section{Spatial Transformer Networks} \label{sec:STN}

In the rapidly emerging field of deep learning among with the explosion of available collected data, deep neural networks have enjoyed huge success in various vision problems. Nevertheless, there had not been a principled way of resolving geometric variations in the given data. 
The recently proposed Spatial Transformer Networks~\cite{jaderberg2015spatial} performs spatial transformations on images or feature maps with a (sub-)differentiable module. It has the effects of reducing geometric variations inside the data and has brought great attention to the deep learning community.

\begin{figure}[t!] \center
\includegraphics[width=\linewidth]{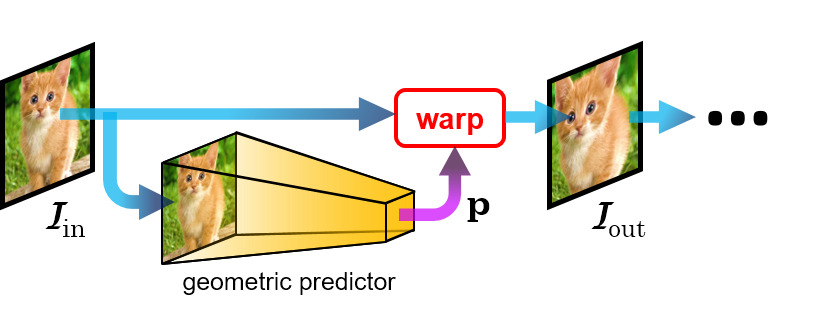}
\caption{Network module of Spatial Transformers~\cite{jaderberg2015spatial}. The blue arrows indicate information passing of appearance, and the purple one indicate that of geometry. The yellow 3D trapezoid denotes the geometric predictor, which contains the learnable parameters.}
\label{fig:ST}
\end{figure}
In the feed-forward sense, a Spatial Transformer warps an image conditioned on the input. This can be mathematically written as
\begin{equation} \label{eq:STN}
\I_{\text{out}}(\0) = \I_{\text{in}}(\p), \;\; \text{where} \;\; \p = f( \I_{\text{in}}(\0) ).
\end{equation}
Here, the nonlinear function $f$ is parametrized as a learnable geometric predictor (termed the localization network in the original paper), which predicts the warp parameters from the input image. 
We note that the ``grid generator'' and the ``sampler'' from the original paper can be combined to be a single warp function.
We can see that for the special case where the geometric predictor consists of a single linear layer, $f$ would consists of a linear regressor $\R$ as well as a bias term $\b$, resulting the geometric predictor in an equivalent form of~\eqref{eq:app-geo-linear}. This insight elegantly links the STN and LK/SDM frameworks together.

Fig.~\ref{fig:ST} shows the basic architecture of STNs.
STNs are of great interest in that transformation predictions can be learned while also showing that grid sampling functions can be (sub-)differentiable, allowing for backpropagation within an end-to-end learning framework. 

Despite the similarities STNs have with classic alignment algorithms, there exist some fundamental drawbacks in comparison to LK/SDM.
For one, it attempts to directly predict the optimal geometric transformation with a single geometric predictor and does not take advantage of the employment of multiple lower-capacity models to achieve more efficient alignment before classification. Although it has been demonstrated that multiple Spatial Transformers can be inserted between feature maps, the effectiveness of such employment has on improving performance is not well-understood.
In addition, we can observe from ~\eqref{eq:STN} that no information of the geometric warp $\p$ is preserved after the output image; this leads to a boundary effect when resampling outside the input source image.
A detailed treatment on this part is provided in Sec.~\ref{sec:geometrypreserve}.

In this work, we aim to improve upon STNs by theoretically connecting it to the LK algorithm.
We show that employing multiple low-capacity models as in LK/SDM for learning spatial transformation within a deep network yields substantial improvement on the subsequent task at hand. 
We further demonstrate the effectiveness of learning a single geometric predictor for recurrent transformation and propose the Inverse Compositional Spatial Transformer Networks (IC-STNs), which exhibit significant improvements over the original STN on various problems.

\section{Inverse Compositional STNs} \label{sec:ICSTN}

\subsection{Geometry Preservation} \label{sec:geometrypreserve}

\begin{figure}[t!] \center
\includegraphics[width=\linewidth]{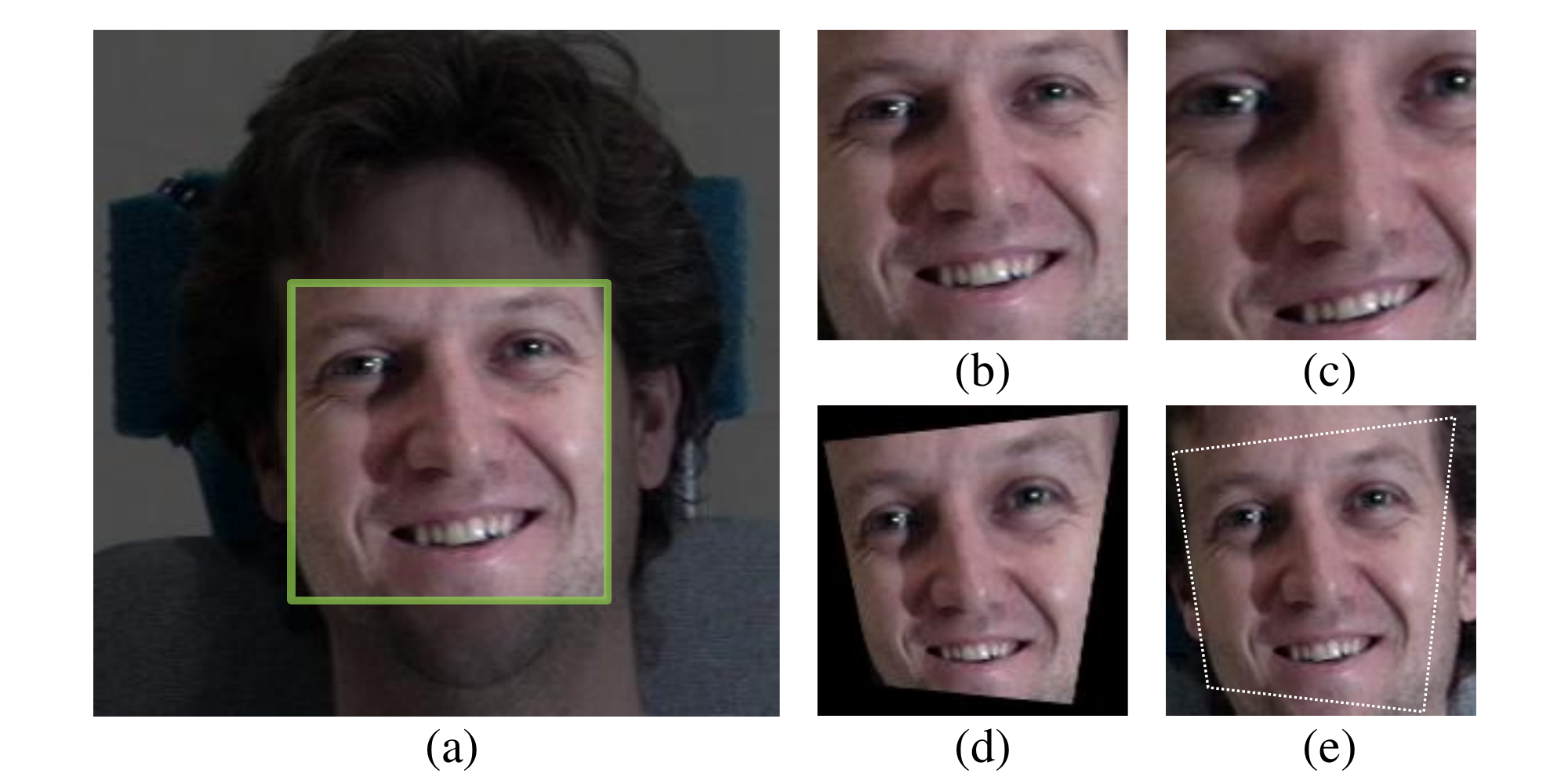}
\caption{Boundary effect of Spatial Transformers on real images.
(a) Original image, where the green box indicates the cropped region.
(b) Cropped image as the input of the Spatial Transformer.
(c) Zoom-in transformation: sampling occurs within the range of the input image.
(d)(e) Zoom-out transformation: discarding the information outside the input image introduces a boundary effect (STNs), while it is not the case with geometry preservation (c-STNs). The white dotted box indicates the warp from the original image. }
\label{fig:boundaryeffect}
\end{figure}

One of the major drawbacks of the original Spatial Transformer architecture (Fig.~\ref{fig:ST}) is that the output image samples only from the cropped input image; pixel information outside the cropped region is discarded,
introducing a boundary effect.
Fig.~\ref{fig:boundaryeffect} illustrates the phenomenon.

We can see from Fig.~\ref{fig:boundaryeffect}(d) that such effect is visible for STNs in zoom-out transformations where pixel information outside the bounding box is required.
This is due to the fact that geometric information is not preserved after the spatial transformations.
In the scenario of iterative alignment, boundary effects are accumulated for each zoom-out transformations.
Although this is less of an issue with images with clean background, this is problematic with real images.

%

\begin{figure}[t!] \center
\includegraphics[width=0.9\linewidth]{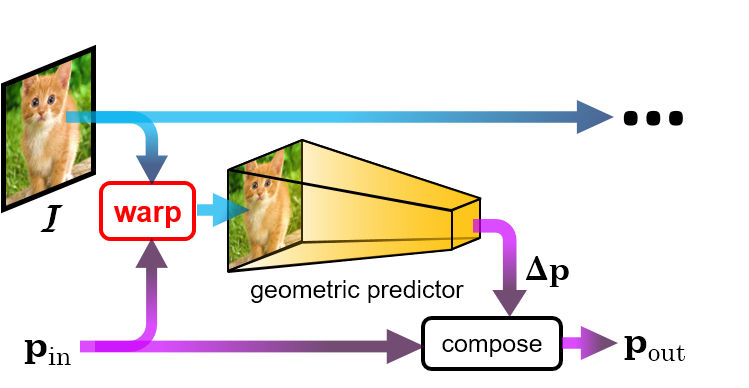}
\caption{A learnable warping module with geometry preserved, termed as c-STNs. The \textit{warp parameters} are passed through the network instead of the \textit{warped images}.}
\label{fig:STwarp}
\end{figure}

\begin{figure*}[t!] \center
\includegraphics[width=0.9\linewidth]{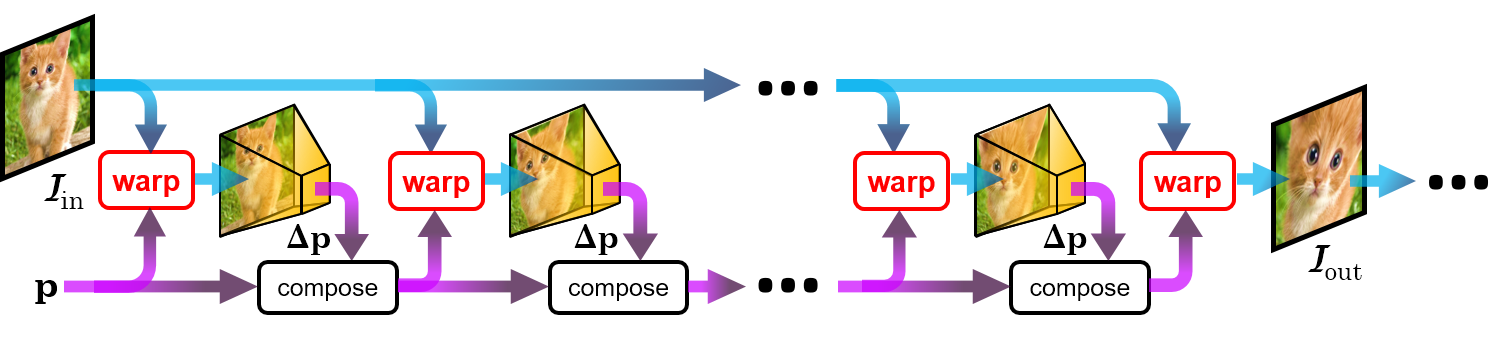}
\caption{Multiple concatenation of c-STNs for an iterative alignment framework.}
\label{fig:multiSTwarp}
\end{figure*}

A series of spatial transformations, however, can be composed and described with exact expressions.
Fig.~\ref{fig:STwarp} illustrates an improved alignment module, which we refer to as compositional STNs (c-STNs). Here, the geometric transformation is also predicted from a geometric predictor, but the \textit{warp parameters} $\p$ are kept track of, composed, and passed through the network instead of the \textit{warped images}. 
It is important to note that if one were to incorporate a cascade of multiple Spatial Transformers, the geometric transformations are implicitly composed through multiple resampling of the images.
We advocate that these transformations are able to be and should be explicitly defined and composed.
Unlike the Spatial Transformer module in Fig.~\ref{fig:ST}, the geometry is preserved in $\p$ instead of being absorbed into the output image.
Furthermore, c-STNs allows repeated concatenation, illustrated in Fig.~\ref{fig:multiSTwarp}, where updates to the warp can be iteratively predicted.
This eliminates the boundary effect because pixel information outside the cropped image is also preserved until the final transformation.

The derivative of warp compositions can also be mathematically expressed in closed forms. Consider the input and output warp parameters $\p_\text{in}$ and $\p_\text{out}$ in Fig.~\ref{fig:STwarp}. Taking the case of affine warps for example, the parameters $\p=[p_1 \;\; p_2 \;\; p_3 \;\; p_4 \;\; p_5 \;\; p_6]^{\top}$ are relatable to transformation matrices in the homogeneous coordinates as
\begin{equation} \label{eq:M(p)}
\M(\p) = 
\begin{bmatrix}
1+p_1 & p_2 & p_3 \\
p_4 & 1+p_5 & p_6 \\
0 & 0 & 1
\end{bmatrix} .
\end{equation}
From the definition of warp composition, the warp parameters are related to the transformation matrices through 
\begin{equation} \label{eq:M(p)compose}
\M(\p_\text{out}) = \M(\deltap) \cdot \M(\p_\text{in}) .
\end{equation}
We can thus derive the derivative to be
\begin{align*} \label{eq:dp/dp}
\frac{\partial \p_\text{out}}{\partial \p_\text{in}} &= \eye +
\begin{bmatrix}
\deltaps_1 & 0 & 0 & \deltaps_2 & 0 & 0 \\
0 & \deltaps_1 & 0 & 0 & \deltaps_2 & 0 \\
0 & 0 & \deltaps_1 & 0 & 0 & \deltaps_2 \\
\deltaps_4 & 0 & 0 & \deltaps_5 & 0 & 0 \\
0 & \deltaps_4 & 0 & 0 & \deltaps_5 & 0 \\
0 & 0 & \deltaps_4 & 0 & 0 & \deltaps_5 \\
\end{bmatrix} \\
\frac{\partial \p_\text{out}}{\partial \deltap} &= \eye +
\begin{bmatrix}
p_{\text{in},1} & p_{\text{in},4} & 0 & 0 & 0 & 0 \\
p_{\text{in},2} & p_{\text{in},5} & 0 & 0 & 0 & 0 \\
p_{\text{in},3} & p_{\text{in},6} & 0 & 0 & 0 & 0 \\
0 & 0 & 0 & p_{\text{in},1} & p_{\text{in},4} & 0 \\
0 & 0 & 0 & p_{\text{in},2} & p_{\text{in},5} & 0 \\
0 & 0 & 0 & p_{\text{in},3} & p_{\text{in},6} & 0 \\
\end{bmatrix} , \numberthis
\end{align*}
where $\eye$ is the identity matrix. This allows the gradients to backpropagate into the geometric predictor.

It is interesting to note that the expression of $\frac{\partial \p_\text{out}}{\partial \p_\text{in}}$ in~\eqref{eq:dp/dp} has a very similar expression as in Residual Networks~\cite{he2015deep,he2016identity}, where the gradients contains the identity matrix $\eye$ and ``residual components''.
This suggests that the warp parameters from c-STNs are generally insensitive to the vanishing gradient phenomenon given the predicted warp parameters $\deltap$ is small, and that it is possible to repeat the warp/composition operation by a large number of times.

We also note that c-STNs are highly analogous to classic alignment algorithms.
If each geometric predictor consists of a single linear layer, \ie the appearance-geometry relationship is assumed to be linearly approximated, then it performs equivalent operations as the compositional LK algorithm.
It is also related to SDM, where heuristic features such as SIFT are extracted before each regression layer.
Therefore, c-STNs can be regarded as a generalization of LK and SDM, differing that the features for predicting the warp updates can be learned from data and incorporated into an end-to-end learning framework.

\subsection{Recurrent Spatial Transformations}

Of all variants of the LK algorithm, the IC form~\cite{baker2004lucas} has a very special property in that the linear regressor remains constant across iterations.
The steepest descent image $\frac{\partial\T(\0)}{\partial\p}$ in~\eqref{eq:LKIC-solution} is independent of the input image and the current estimate of $\p$; therefore, it is only needed to be computed once.
In terms of model capacity, IC-LK further reduces the necessary learnable parameters compared to canonical LK, for the same regressor can be applied repeatedly and converges provided a good initialization. The main difference from canonical LK and IC-LK lies in that the warp update $\deltap$ should be compositionally applied in the inverse form.
We redirect the readers to~\cite{baker2004lucas} for a full treatment of IC-LK, which is out of scope of this paper.

\begin{figure}[t!] \center
\includegraphics[width=\linewidth]{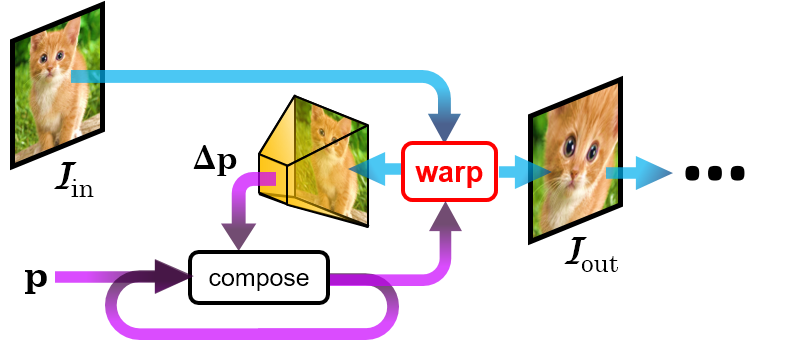}
\caption{Illustration of the proposed Inverse Compositional Spatial Transformer Network (IC-STN). The same geometric predictor is learned to predict recurrent spatial transformations that are composed together to warp the input image.}
\label{fig:IC-STN}
\end{figure}

This inspires us to propose the Inverse Compositional Spatial Transformer Network (IC-STN). Fig.~\ref{fig:IC-STN} illustrates the recurrent module of IC-STN: the warp parameters $\p$ is iteratively updated by $\deltap$, which is predicted from the current warped image with the same geometric predictors. This allows one to recurrently predict spatial transformations on the input image. It is possible due to the close spatial proximity of pixel intensities within natural images: there exists high correlation between pixels in close distances.

In the IC-LK algorithm, the predicted warp parameters are inversely composed. Since the IC-STN geometric predictor is optimized in an end-to-end learning framework, we can absorb the inversion operation into the geometric predictor without explicitly defining it; in other words, IC-STNs are able to directly predict the inverse parameters. 
In our experiments, we find that there is negligible difference to explicitly perform an additional inverse operation on the predicted forward parameters, and that implicitly predicting the inverse parameters fits more elegantly in an end-to-end learning framework using backpropagation.
We name our proposed method Inverse Compositional nevertheless as IC-LK is where our inspirations are drawn from.

In practice, IC-STNs can be trained by unfolding the architecture in Fig.~\ref{fig:IC-STN} multiple times into the form of c-STNs (Fig.~\ref{fig:multiSTwarp}), sharing the learnable parameters across all geometric predictors, and backpropagating the gradients as described in Sec.~\ref{sec:geometrypreserve}. This results in a single effective geometric predictor that can be applied multiple times before performing the final warp operation that suits subsequent tasks such as classification.


\section{Experiments} \label{sec:experiments}

\subsection{Planar Image Alignment} \label{sec:planar}

\begin{figure}[t!] \center
\includegraphics[width=\linewidth]{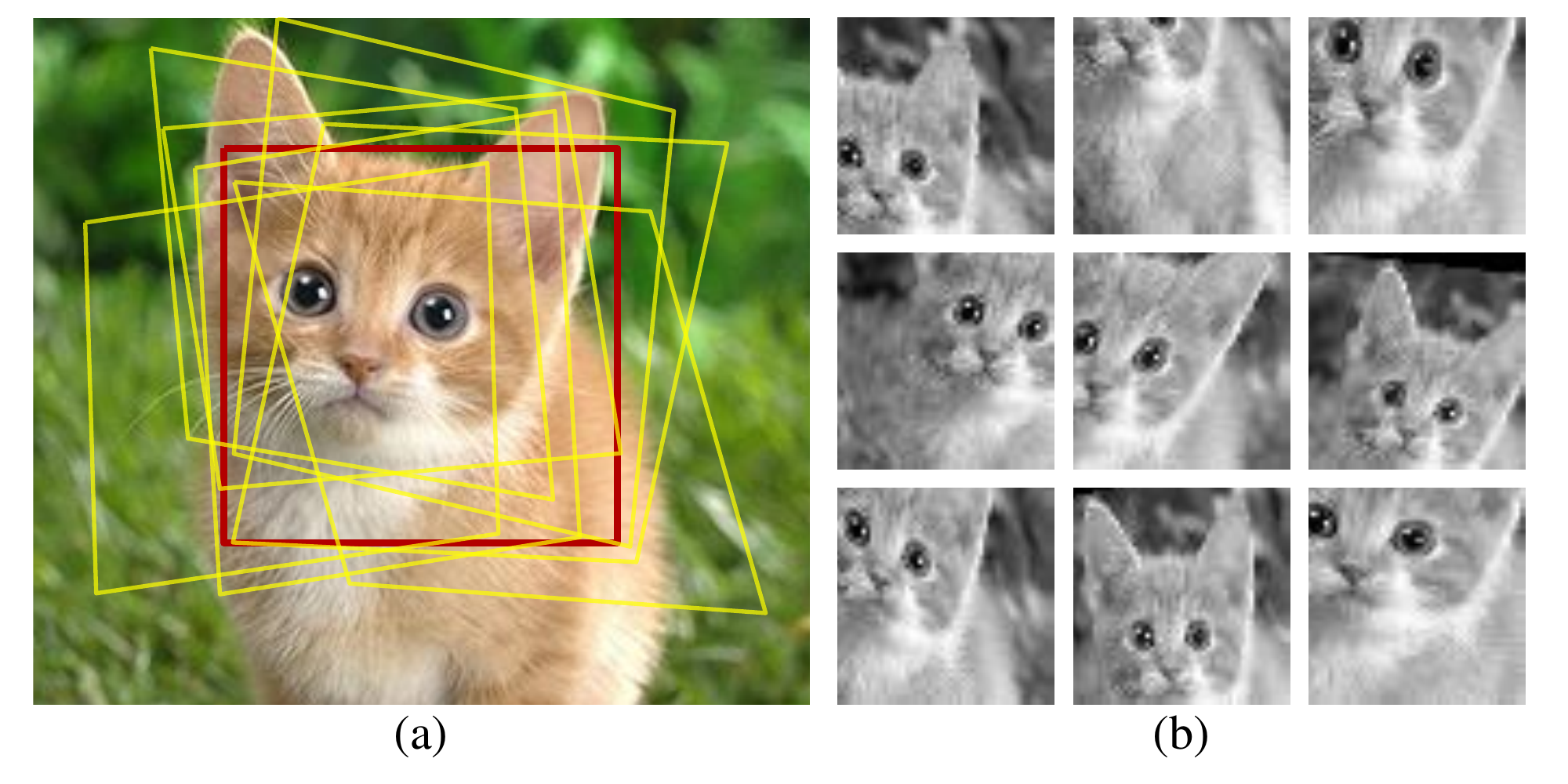}
\caption{Visualization of the image and perturbed training samples for the planar image alignment experiment. (a) Original image, where the red box indicates the ground-truth warp and the yellow boxes indicate example generated warps. (b) Examples of the perturbed images (affine warps with $\sigma=7.5$ in this case).}
\label{fig:planar}
\end{figure}

To start with, we explore the efficacy of IC-STN for planar alignment of a single image. We took an example image from the Caffe library~\cite{jia2014caffe} and generated perturbed images with affine warps around the hand-labeled ground truth, shown in Fig.~\ref{fig:planar}. We used image samples of size 50$\times$ 50 pixels. The perturbed boxes are generated by adding i.i.d. Gaussian noise of standard deviation $\sigma$ (in pixels) to the four corners of the ground-truth box plus an additional translational noise from the same Gaussian distribution, and finally fitting the box to the initial warp parameters $\p$.


\begin{table}[t!] \center
\begin{tabular}{c|c|c|c|c}
Model & $\sigma=2.5$ & $\sigma=5$ & $\sigma=7.5$ & $\sigma=10$ \\ \hline
c-STN-1 & 2.699 & 5.576 & 9.491 & 9.218 \\
IC-STN-2 & 0.615 & 2.268 & 5.283 & 5.502 \\
IC-STN-3 & 0.434 & 1.092 & 2.877 & 3.020 \\
IC-STN-4 & 0.292 & 0.481 & 1.476 & 2.287 \\
IC-STN-6 & 0.027 & 0.125 & 0.245 & 1.305 \\
\end{tabular}
\caption{Test error for the planar image alignment experiment under different extents of initial perturbations. The number following the model names indicate the number of warp operations unfolded from IC-STN during training.}
\label{table:planar}
\end{table}

To demonstrate the effectiveness of iterative alignment under different amount of noise, we consider IC-STNs that consist of a single learnable linear layer with different numbers of learned recurrent transformations.
We optimize all networks in terms of $L_2$ error between warp parameters with stochastic gradient descent and a batch size of 100 perturbed training samples generated on the fly.

The test error is illustrated in Table~\ref{table:planar}. We see from c-STN-1 (which is equivalent to IC-STN-1 with only one warp operation unfolded) that a single geometric warp predictor has limited ability to directly predict the optimal geometric transformation. Reusing the geometric predictor to incorporating multiple spatial transformations yields better alignment performance given the same model capacity.

\begin{figure}[t!] \center
\includegraphics[width=0.8\linewidth]{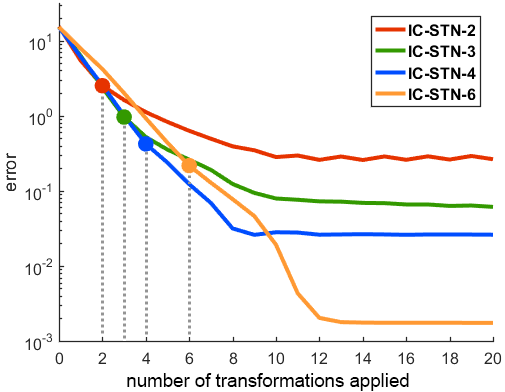}
\caption{Evaluation on trained IC-STNs, where the dot on each curve corresponds to the number of recurrent transformations unfolded during training.}
\label{fig:planar-recur}
\end{figure}

Fig.~\ref{fig:planar-recur} shows the test error over the number of warp operations applied to the learned alignment module. We can see that even when the recurrent spatial transformation is applied more times than trained with, the error continues to decrease until some of point of saturation, which typically does not hold true for classical recurrent neural networks. This implies that IC-STN is able to capture the correlation between appearance and geometry to perform gradient descent on a learned cost surface for successful alignment.

\subsection{MNIST Classification} \label{sec:MNIST}


In this section, we demonstrate how IC-STNs can be utilized in joint alignment/classfication tasks. We choose the MNIST handwritten digit dataset~\cite{lecun1998mnist}, and we use a homography warp noise model to perturb the four corners of the image and translate them with Gaussian noise, both with a standard deviation of 3.5 pixels.
We train all networks for 200K iterations with a batch size of 100 perturbed samples generated on the fly.
We choose a constant learning rate of 0.01 for the classification subnetworks and 0.0001 for the geometric predictors as we find the geometric predictor sensitive to large changes.
We evaluate the classification accuracy on the test set using the same warp noise model.

\begin{table*}[t!] \center
\begin{tabular}{c|c|c|l}
Model & Test error & Capacity & \multicolumn{1}{c}{Architecture} \\ \hline
CNN(a) & 6.597 \% & 39079 & \CNNcolor conv(3$\times$3, 3)-conv(3$\times$3, 6)-P-conv(3$\times$3, 9)-conv(3$\times$3, 12)-FC(48)-FC(10) \\
STN(a) & 4.944 \% & 39048 & \STcolor [ conv(7$\times$7, 4)-conv(7$\times$7, 8)-P-FC(48)-FC(8) ]\textbf{$\times$1} \black $\to$ \CNNcolor conv(9$\times$9, 3)-FC(10) \\
c-STN-1(a) & 3.687 \% & 39048 & \STcolor [ conv(7$\times$7, 4)-conv(7$\times$7, 8)-P-FC(48)-FC(8) ]\textbf{$\times$1} \black $\to$ \CNNcolor conv(9$\times$9, 3)-FC(10) \\
c-STN-2(a) & 2.060 \% & 38528 & \STcolor [ conv(9$\times$9, 4)-FC(8) ]\textbf{$\times$2} \black $\to$ \CNNcolor conv(9$\times$9, 3)-FC(10) \\
c-STN-4(a) & 1.476 \% & 37376 & \STcolor [ FC(8) ]\textbf{$\times$4} \black $\to$ \CNNcolor conv(9$\times$9, 3)-FC(10) \\
IC-STN-2(a) & \textbf{1.905} \% & 39048 & \STcolor [ conv(7$\times$7, 4)-conv(7$\times$7, 8)-P-FC(48)-FC(8) ]\textbf{$\times$2} \black $\to$ \CNNcolor conv(9$\times$9, 3)-FC(10) \\
IC-STN-4(a) & \textbf{1.230} \% & 39048 & \STcolor [ conv(7$\times$7, 4)-conv(7$\times$7, 8)-P-FC(48)-FC(8) ]\textbf{$\times$4} \black $\to$ \CNNcolor conv(9$\times$9, 3)-FC(10) \\ \hline
CNN(b) & 19.065 \% & 19610 & \CNNcolor conv(9$\times$9, 2)-conv(9$\times$9, 4)-FC(32)-FC(10) \\
STN(b) & 9.325 \% & 18536 & \STcolor [ FC(8) ]\textbf{$\times$1} \black $\to$ \CNNcolor conv(9$\times$9, 3)-FC(10) \\
c-STN-1(b) & 8.545 \% & 18536 & \STcolor [ FC(8) ]\textbf{$\times$1} \black $\to$ \CNNcolor conv(9$\times$9, 3)-FC(10) \\
IC-STN-2(b) & \textbf{3.717} \% & 18536 & \STcolor [ FC(8) ]\textbf{$\times$2} \black $\to$ \CNNcolor conv(9$\times$9, 3)-FC(10) \\
IC-STN-4(b) & \textbf{1.703} \% & 18536 & \STcolor [ FC(8) ]\textbf{$\times$4} \black $\to$ \CNNcolor conv(9$\times$9, 3)-FC(10) \\

\end{tabular}
\caption{Classification error on the perturbed MNIST test set.
The non-recurrent networks have similar numbers of layers and learnable parameters but different numbers of warp operations (bold-faced).
The filter dimensions are shown in parentheses, where those of the geometric predictor(s) are in green and those of the subsequent classification network are in blue (P denotes a 2$\times$2 max-pooling operation).
Best viewed in color.}
\label{table:MNISTresult}
\end{table*}

We compare IC-STN to several network architectures, including a baseline CNN with no spatial transformations, the original STN from Jaderberg \etal, and c-STNs. All networks with spatial transformations employ the same classification network.
The results as well as the architectural details are listed in Table~\ref{table:MNISTresult}.
We can see that classical CNNs do not handle large spatial variations efficiently with data augmentation. In the case where the digits may be occluded, however, trading off capacity for a single deep predictor of geometric transformation also results in poor performance.
Incorporating multiple transformers lead to a significant improvement in classification accuracy; further comparing c-STN-4(a) and IC-STN-4(b), we see that IC-STNs are able to trade little accuracy off for a large reduction of capacity compared to its non-recurrent counterpart. 

\begin{figure*}[t!] \center
\includegraphics[width=\linewidth]{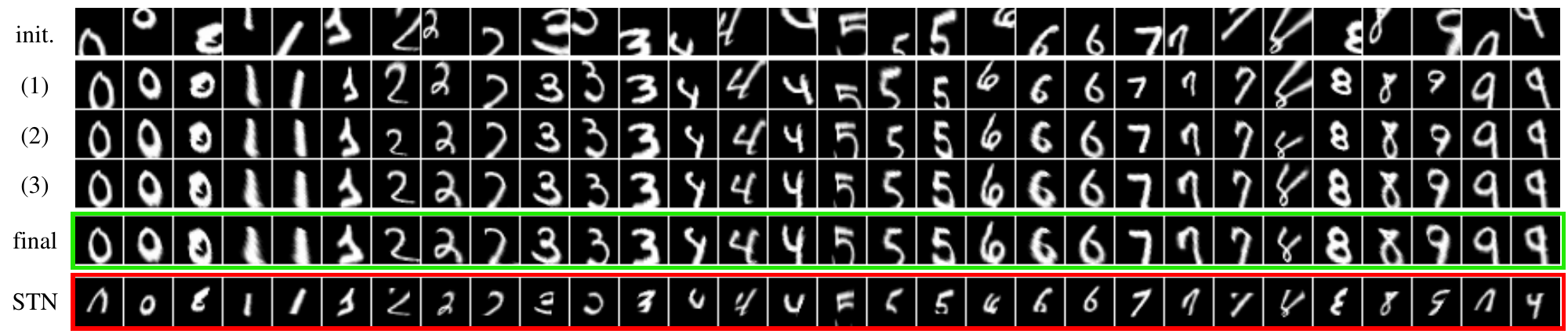}
\caption{Sample alignment results of IC-STN-4(a) on the MNIST test set with homography warp perturbations. The first row of each column shows the initial perturbation; the middle three rows illustrates the alignment process (iterations 1 to 3); the second last row shows the final alignment before feeding into the classification network. The last row shows the alignment from the original STN: the cropped digits are the results of the boundary effect.}
\label{fig:MNISTalign}
\end{figure*}

Fig.~\ref{fig:MNISTalign} shows how IC-STNs learns alignment for classification. In many cases where the handwritten digits are occluded, IC-STN is able to automatically warp the image and reveal the occluded information from the original image. There also exists smooth transitions during the alignment, which confirms with the recurrent spatial transformation concept IC-STN learns. Furthermore, we see that the outcome of the original STN becomes cropped digits due to the boundary effect described in Sec.~\ref{sec:geometrypreserve}.

\begin{figure}[t!] \center
\includegraphics[width=\linewidth]{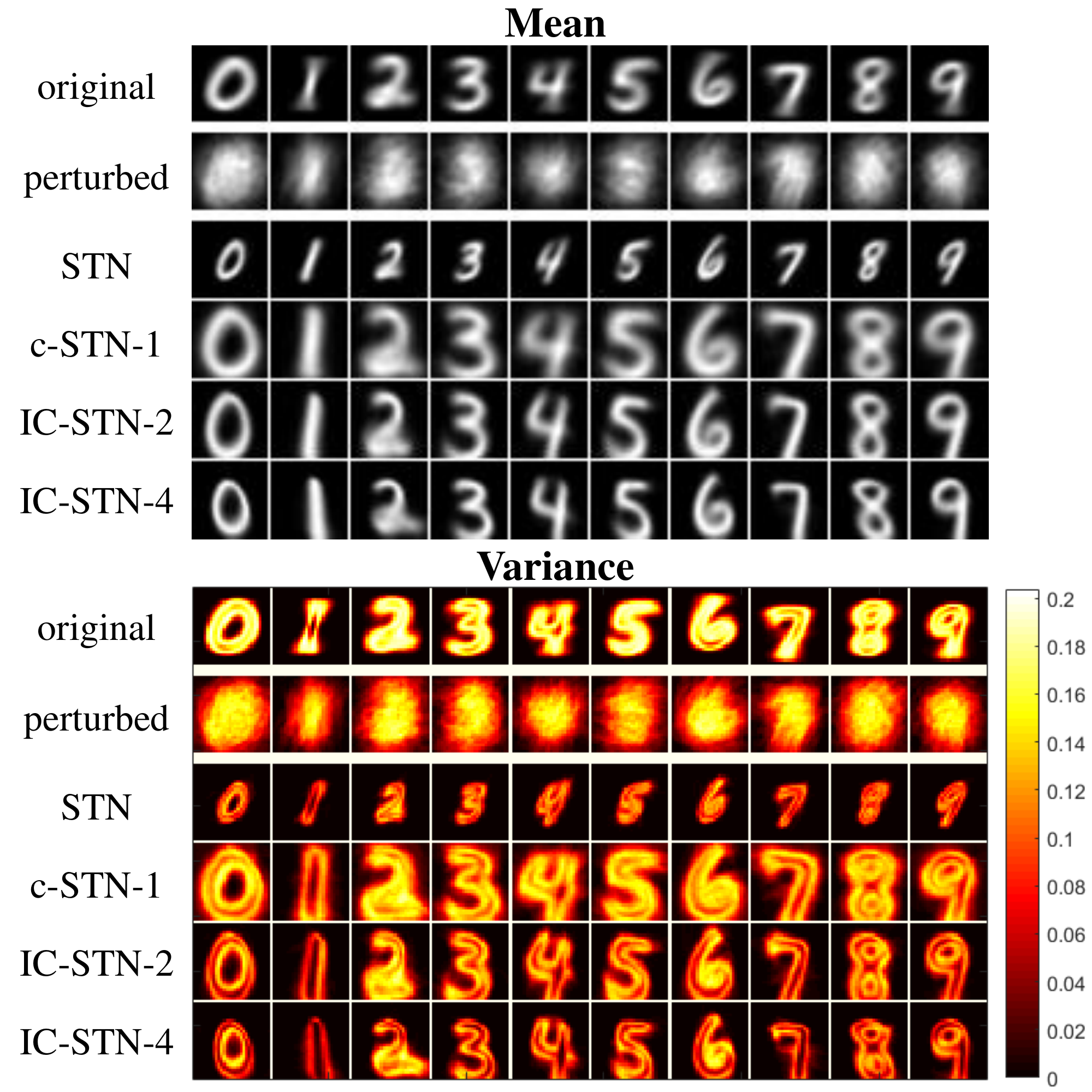}
\caption{Mean/variance of the aligned appearances from the 10 classes of the test set (homography perturbations).}
\label{fig:MNISTmeanvar}
\end{figure}

We also visualize the overall final alignment performance by taking the mean and variance on the test set appearance before classification, shown in Fig.~\ref{fig:MNISTmeanvar}. 
The mean/variance results of the original STN becomes a down-scaled version of the original digits, reducing information necessary for better classification.
From c-STN-1, we see that a single geometric predictor is poor in directly predicting geometric transformations. The variance among all aligned samples is dramatically decreased when more warp operations are introduced in IC-STN. These results support the fact that elimination of spatial variations within data is crucial to boosting the performance of subsequent tasks.

\subsection{Traffic Sign Classification}


Here, we show how IC-STNs can be applied to real-world classification problems such as traffic sign recognition. We evaluate our proposed method with the German Traffic Sign Recognition Benchmark~\cite{stallkamp2011german}, which consists of 39,209 training and 12,630 test images from 43 classes taken under various conditions. We consider this as a challenging task since many of the images are taken with motion blurs and/or of resolution as low as 15$\times$15 pixels.
We rescale all images and generate perturbed samples of size 36$\times$36 pixels with the same homography warp noise model described in Sec.~\ref{sec:MNIST}.
The learning rate is set to be 0.001 for the classification subnetworks and 0.00001 for the geometric predictors.

\begin{table*}[t!] \center
\begin{tabular}{c|c|c|l}
Model & Test error & Capacity & \multicolumn{1}{c}{Architecture} \\ \hline
CNN & 8.287 \% & 200207 & \CNNcolor conv(7$\times$7, 6)-conv(7$\times$7, 12)-P-conv(7$\times$7, 24)-FC(200)-FC(43) \\
STN & 6.495 \% & 197343 & \STcolor [ conv(7$\times$7, 6)-conv(7$\times$7, 24)-FC(8) ]\textbf{$\times$1} \black $\to$ \CNNcolor conv(7$\times$7, 6)-conv(7$\times$7, 12)-P-FC(43) \\
c-STN-1 & 5.011 \% & 197343 & \STcolor [ conv(7$\times$7, 6)-conv(7$\times$7, 24)-FC(8) ]\textbf{$\times$1} \black $\to$ \CNNcolor conv(7$\times$7, 6)-conv(7$\times$7, 12)-P-FC(43) \\
IC-STN-2 & 4.122 \% & 197343 & \STcolor [ conv(7$\times$7, 6)-conv(7$\times$7, 24)-FC(8) ]\textbf{$\times$2} \black $\to$ \CNNcolor conv(7$\times$7, 6)-conv(7$\times$7, 12)-P-FC(43) \\
IC-STN-4 & \textbf{3.184} \% & 197343 & \STcolor [ conv(7$\times$7, 6)-conv(7$\times$7, 24)-FC(8) ]\textbf{$\times$4} \black $\to$ \CNNcolor conv(7$\times$7, 6)-conv(7$\times$7, 12)-P-FC(43) \\

\end{tabular}
\caption{Classification error on the perturbed GTSRB test set.
The architectural descriptions follow that in Table~\ref{table:MNISTresult}.}
\label{table:GTSRBresult}
\end{table*}

We set the controlled model capacities to around 200K learnable parameters and perform similar comparisons to the MNIST experiment. Table~\ref{table:GTSRBresult} shows the classification error on the perturbed GTSRB test set. Once again, we see a considerable amount of classification improvement of IC-STN from learning to reuse the same geometric predictor.
 
\begin{figure}[t!] \center
\includegraphics[width=\linewidth]{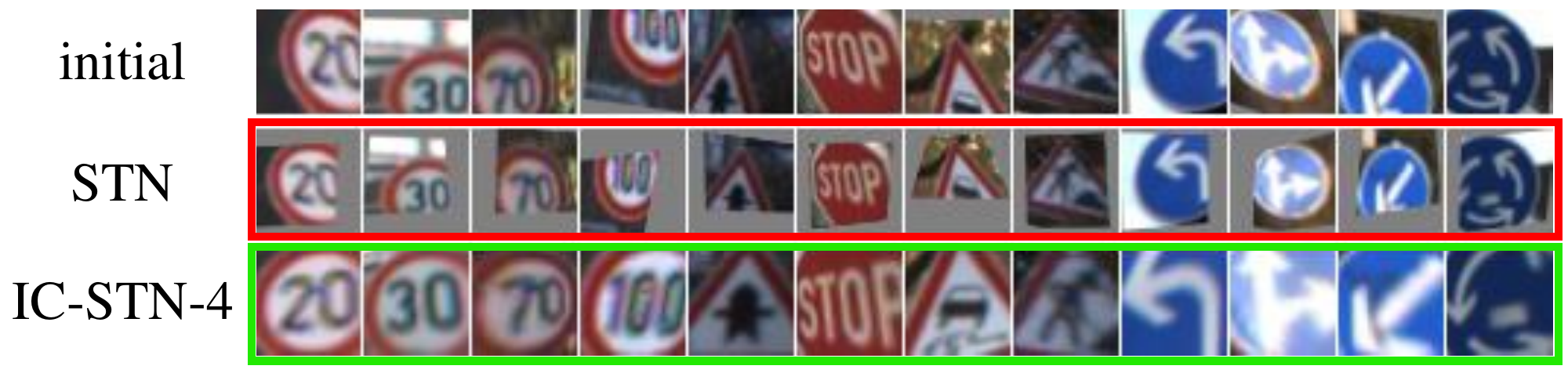}
\caption{Sample alignment results of IC-STN-4 on the GTSRB test set in comparison to the original STN.}
\label{fig:GTSRBalign}
\end{figure}

\begin{figure}[t!] \center
\includegraphics[width=\linewidth]{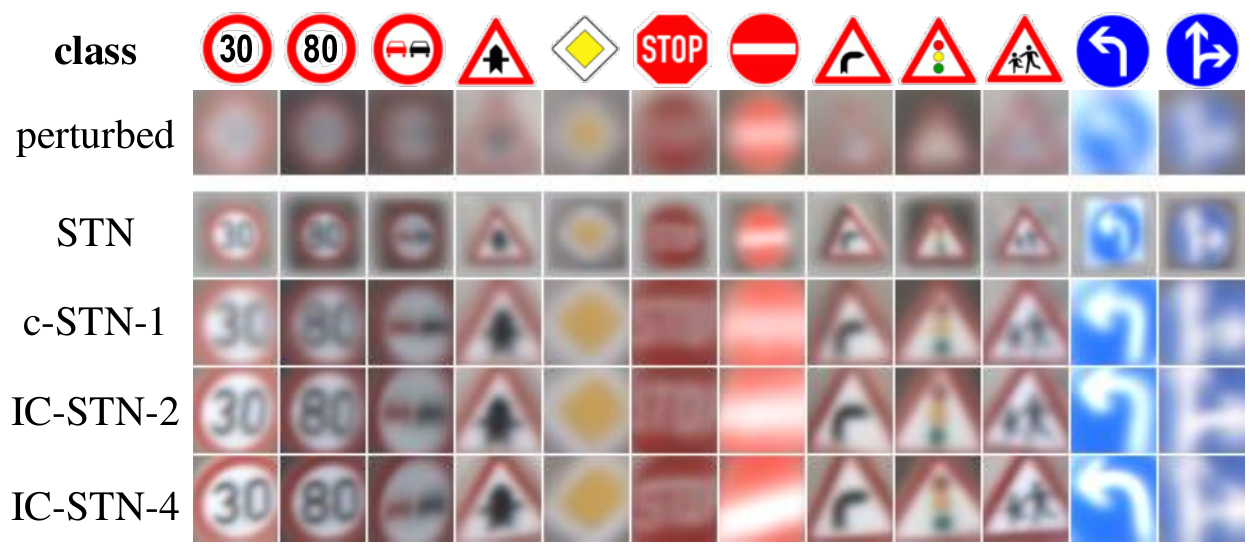}
\caption{Mean aligned appearances for classification from sampled classes of the GTSRB test set.}
\label{fig:GTSRBmean}
\end{figure}

Fig.~\ref{fig:GTSRBalign} compares the aligned images from IC-STN and the original STN before the classification networks. Again, IC-STNs are able to recover occluded appearances from the input image. Although STN still attempts to center the perturbed images, the missing information from occlusion degrades its subsequent classification performance.

We also visualize the aligned mean appearances from each network in Fig.~\ref{fig:GTSRBmean}, and it can be observed that the mean appearance of IC-STN becomes sharper as the number of warp operations increase, once again indicating that good alignment is crucial to the subsequent target tasks.
It is also interesting to note that not all traffic signs are aligned to be fit exactly inside the bounding boxes, \eg the networks finds the optimal alignment for stop signs to be zoomed-in images while excluding the background information outside the octagonal shapes.
This suggests that in certain cases, only the pixel information inside the sign shapes are necessary to achieve good alignment for classification.

\section{Conclusion} \label{sec:conclusion}

In this paper, we theoretically connect the core idea of the Lucas \& Kanade algorithm with Spatial Transformer Networks. We show that geometric variations within data can be eliminated more efficiently through multiple spatial transformations within an alignment framework. We propose Inverse Compositional Spatial Transformer Networks for predicting recurrent spatial transformations and demonstrate superior alignment and classification results compared to baseline CNNs and the original STN.

{\small
\bibliographystyle{ieee}
\bibliography{reference}
}

\end{document}

%% file: notation.tex
\newcommand{\fig}[1]{Figure \ref{Fig:#1}}
\newcommand{\eqn}[1]{Eqn. \ref{eqn:#1}}
%
\newcommand{\eucsqnorm}[1]{\left\|{#1}\right\|_2^2}
\newcommand{\frobnorm}[1]{\left\|{#1}\right\|_F^2}

\newcommand{\I}{\mathcal{I}}
\newcommand{\T}{\mathcal{T}}
\newcommand{\p}{\mathbf{p}}
\newcommand{\0}{\mathbf{0}}
\newcommand{\deltap}{\Delta \p}
\newcommand{\deltaps}{\Delta p}
\newcommand{\Ip}{\I(\p)}
\newcommand{\TO}{\T(\0)}
\newcommand{\y}{\mathbf{y}}
\newcommand{\x}{\mathbf{x}}
\newcommand{\deltax}{\Delta \x}
\newcommand{\dTOdp}{\frac{\partial \T(\0)}{\partial \p}}
\newcommand{\dxdp}{\frac{\partial \mathcal{W}(\x; \0)}{\partial \p^{\top} }}
\newcommand{\R}{\mathbf{R}}
\newcommand{\W}{\mathcal{W}}
\renewcommand{\b}{\mathbf{b}}

\newcommand{\eye}{\mathbf{I}}
\newcommand{\Real}{\mathbb{R}}
\newcommand{\X}{\mathbf{X}}
\newcommand{\Y}{\mathbf{Y}}
\newcommand{\M}{\mathbf{M}}
%
%
\newcommand{\diag}{\mbox{diag}}
\newcommand{\kron}{\otimes}
\newcommand{\N}{\mathcal{N}}
\renewcommand{\vec}{\mbox{vec}}
\newcommand{\st}{\mbox{s.t.}}
\newcommand{\grad}{\nabla}
\newcommand{\qsection}[1]{\vspace{4mm} \noindent \textbf{#1:}}
%
%
%